\documentclass[conference]{IEEEtran}
\IEEEoverridecommandlockouts
\usepackage{hyperref}
\usepackage{graphicx}
\usepackage{subfigure}
\usepackage{setspace}
\usepackage{xcolor}
\usepackage{multicol}
\DeclareUnicodeCharacter{2212}{-}
\DeclareUnicodeCharacter{2009}{-}
\usepackage{amsmath}
\usepackage{amssymb}
\usepackage{float}

\usepackage{caption}
\usepackage{cite}
\usepackage{amsmath,amssymb,amsfonts}
\usepackage{graphicx}
\usepackage{array}
\usepackage{booktabs}
\usepackage{caption}
\usepackage{ragged2e}
\usepackage{makecell} 
\usepackage{adjustbox} 
\usepackage{anyfontsize} 
\usepackage{booktabs}
\usepackage{array}
\usepackage{tabularx} 
\usepackage{array}
\usepackage{graphicx}
\usepackage{textcomp}
\usepackage{xcolor}
\usepackage{multirow}
\usepackage{booktabs}
\usepackage{graphicx}
\usepackage{multirow}
\usepackage{threeparttable}

\usepackage{xcolor}
\usepackage{cite}
\usepackage{amsmath,amssymb,amsfonts}
\usepackage{algorithmic}
\usepackage{graphicx}
\usepackage{textcomp}
\usepackage{xcolor}
\def\BibTeX{{\rm B\kern-.05em{\sc i\kern-.025em b}\kern-.08em
    T\kern-.1667em\lower.7ex\hbox{E}\kern-.125emX}}
    
\begin{document}

\title{BanglaDialecto: An End-to-End AI-Powered Regional Speech Standardization}

\author{
    \IEEEauthorblockN{Md. Nazmus Sadat Samin\textsuperscript{*}}
    \IEEEauthorblockA{\textit{Apurba-NSU R\&D Lab, ECE} \\
    \textit{North South University}\\
    Dhaka, Bangladesh \\
    nazmus.samin@northsouth.edu}
    \and
    \IEEEauthorblockN{Jawad Ibn Ahad\textsuperscript{*}}
    \IEEEauthorblockA{\textit{Apurba-NSU R\&D Lab, ECE} \\
    \textit{North South University}\\
    Dhaka, Bangladesh \\
    jawad.ibn@northsouth.edu}
    \and
    \IEEEauthorblockN{Tanjila Ahmed Medha}
    \IEEEauthorblockA{\textit{Apurba-NSU R\&D Lab, ECE} \\
    \textit{North South University}\\
    Dhaka, Bangladesh \\
    tanjila.medha@northsouth.edu}
    \and
    \IEEEauthorblockN{Fuad Rahman}
    \IEEEauthorblockA{\textit{Apurba Technologies}\\
    Sunnyvale, CA 94085, USA \\
    fuad@apurbatech.com}
    \and 
    \IEEEauthorblockN{Mohammad Ruhul Amin}
    \IEEEauthorblockA{\textit{Computer and Information Science} \\
    \textit{Fordham University}\\
    New York, USA \\
    mamin17@fordham.edu}
    \and
    \IEEEauthorblockN{Nabeel Mohammed}
    \IEEEauthorblockA{\textit{Apurba-NSU R\&D Lab, ECE} \\
    \textit{North South University}\\
    Dhaka, Bangladesh \\
    nabeel.mohammed@northsouth.edu}
    \and 
    \IEEEauthorblockN{Shafin Rahman}
    \IEEEauthorblockA{\textit{Apurba-NSU R\&D Lab, ECE} \\
    \textit{North South University}\\
    Dhaka, Bangladesh \\
    shafin.rahman@northsouth.edu}
    \thanks{\textsuperscript{*}Equal contribution}
}

\maketitle

\begin{abstract}
This study focuses on recognizing Bangladeshi dialects and converting diverse Bengali accents into standardized formal Bengali speech. Dialects, often referred to as regional languages, are distinctive variations of a language spoken in a particular location and are identified by their phonetics, pronunciations, and lexicon. Subtle changes in pronunciation and intonation are also influenced by geographic location, educational attainment, and socioeconomic status. Dialect standardization is needed to ensure effective communication, educational consistency, access to technology, economic opportunities, and the preservation of linguistic resources while respecting cultural diversity. Being the fifth most spoken language with around 55 distinct dialects spoken by 160 million people, addressing Bangla dialects is crucial for developing inclusive communication tools. However, limited research exists due to a lack of comprehensive datasets and the challenges of handling diverse dialects. With the advancement in multilingual Large Language Models (mLLMs), emerging possibilities have been created to address the challenges of dialectal Automated Speech Recognition (ASR) and Machine Translation (MT). This study presents an end-to-end pipeline for converting dialectal Noakhali speech to standard Bangla speech. This investigation includes constructing a large-scale diverse dataset with dialectal speech signals that tailored the fine-tuning process in ASR and LLM for transcribing the dialect speech to dialect text and translating the dialect text to standard Bangla text. Our experiments demonstrated that fine-tuning the Whisper ASR model achieved a CER of 0.8\% and WER of 1.5\%, while the BanglaT5 model attained a BLEU score of 41.6\% for dialect-to-standard text translation. We completed our end-to-end pipeline for dialect standardization by utilizing AlignTTS, a text-to-speech (TTS) model. With potential applications across different dialects, this research lays the groundwork for future research into Bangla dialect standardization. 

\end{abstract}


\section{Introduction}

\begin{figure}[!t]
    \centering
    \includegraphics[scale=0.155]{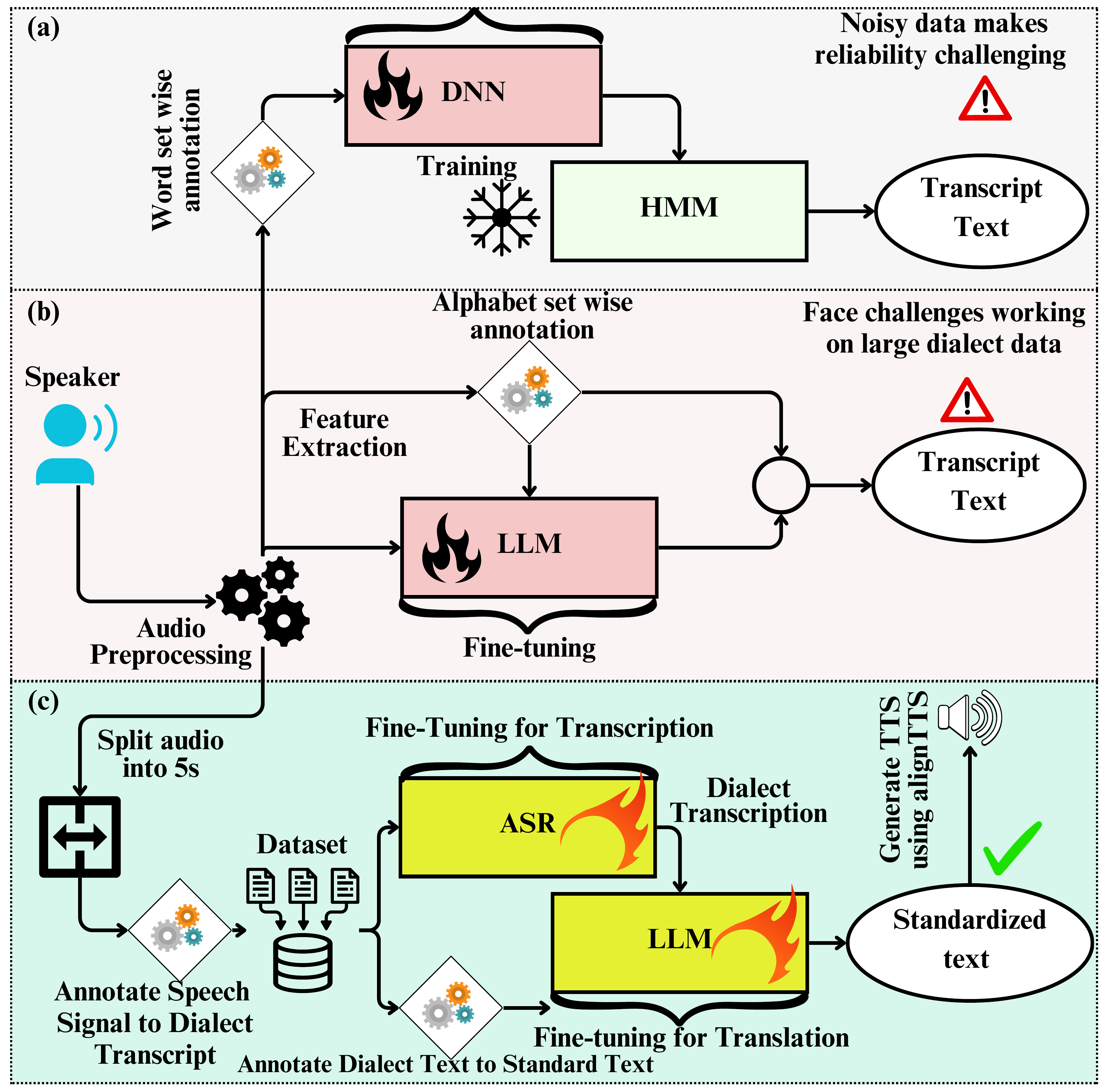}
    \caption{(a) Typical Deep Neural Network (DNN) based implementation of ASR that uses word-based annotation combining HMM is used by~\cite{al2019continuous, khan2018assessing,samin2021deep}, (b) Recent advancement of LLMs let researchers to get involved in investigation towards the LLMs capability of handling speech signal. The mLLM-based approach has been applied in \cite{gudepu2020whisper, pratama2024analysis, phung2024enhancing} using feature extraction and alphabet-wise mapping. Existing methods often fall short of processing big speech signal data, specifically with dialect speech signals, due to limitations of data availability and resources. On the other hand, end-to-end frameworks are less explored as per the literature. (c) We introduce a novel approach involving fine-tuning ASR and mLLMs with a large-scale low-resource Bangla dialect speech signal dataset. There are two parts, one is the dialect transcript from the dialect speech signal, which will be performed by the multilingual ASR model and then LLM will translate and standardize the ASR models' predicted dialect text into standard Bangla text. Our approach includes reliable preprocessing techniques to handle large-scale speech signals.}
    \label{fig:figure_1}
\end{figure}
A dialect, also referred to as a regional language, is a distinct version of a language spoken in a specific geographical area, characterized by unique phonetics, pronunciation, and vocabulary \cite{rahman2024phonological}. Children naturally acquire it as their first language from their caregivers, without formal grammar instruction. Societal factors such as socio-economic background, education levels, and geographic location further contribute to subtle variations in pronunciation and intonation \cite{hamed2023social}. Bangla, the language spoken by approximately 160 million people across 64 districts in Bangladesh, is the fifth most spoken language in the world \cite{siddique2021english}. Interestingly, there are around 55 distinct dialects of Bangla \cite{faria2023vashantor}. Addressing these language variations is crucial for creating inclusive and effective communication tools for Bengali speakers. This framework could potentially be expanded to accommodate dialects from other regions globally. Despite the cultural significance of these dialects, research on Bangla regional speech remains limited, primarily due to the absence of a comprehensive database and the complexities associated with handling such extensive and diverse datasets. Dialectal differences often hinder people's access to essential services, effective communication, medical treatment, and education and impede their social and professional interactions. The advancement of automatic speech recognition (ASR) technology is imperative in bridging the gap between human language and machine comprehension \cite{yu2016automatic}. This study aims to tackle the challenges posed by Bengali regional accents in voice recognition, particularly focusing on addressing the complexities of mining big data for this purpose. The project endeavors to remove barriers to effective communication and improve the accuracy and inclusivity of automated speech-to-text systems by developing a robust system capable of recognizing and processing a wide range of regional accents.

A reliable ASR system is essential for local communities in regions with diverse accents. Traditionally, researchers have heavily relied on conventional voice recognition methods. \emph{\textbf{(a) DNN-based ASR:}} Many studies on Bangla ASR have employed the deep neural network (DNN) approach \cite{tausif2018deep}. Researchers have attempted to enhance ASR performance by integrating DNN with other models such as GMM and HMM \cite{al2019continuous}. However, there has been no recent effort to apply these methods to identify Bengali dialects due to the requirement of substantial data for accurate speech recognition. The DNN approach has exhibited great potential in machine translation (MT) over the past decade. DNNs are utilized in MT to learn syntactic and semantic representations of language, thereby improving accuracy and resilience to input variations through word embedding in a continuous space\cite{zhang2015deep}. \emph{\textbf{(b) Large Language models (LLMs) in speech-related tasks:}} The field of natural language processing (NLP) has undergone a significant transformation due to pre-trained language models (PLMs). \textcolor{black}{The development of LLM models such as GPT, BERT \cite{devlin2018bert}, and ASR models like Whisper \cite{radford2023robust} }represents exciting advancements. These models have broad applicability in speech-related tasks including question-and-answer systems, speech recognition, text generation, MT, and natural language generation. OpenAI's whisper ASR model has been extensively utilized in low-resource language speech recognition systems \cite{pratama2024analysis} and can be adapted for speech-related tasks using various fine-tuning methods \cite{liu2024exploration}.

There has been no action taken to explore the Bangla dialect ASR system using mLLMs or Whisper ASR due to the lack of a sufficient dataset. A large dataset of the local language is required to train these models, which is currently unavailable. LLMs have been widely utilized in MT tasks due to their precise multilingual translation capabilities \cite{zhu2023multilingual}. Some Bangladeshi scholars have started to incorporate mLLMs in MT for the Bengali language \cite{faria2023vashantor}. However, existing studies have employed these approaches for individual tasks such as Automatic Speech Recognition (ASR), Machine Translation (MT), or Text-to-speech (TTS). Integrating these three speech-related tasks into a single pipeline would greatly benefit the local speakers. Considering the Bengali dialect, resources are extremely limited. Additionally, the variations in phonemes and the availability of audio create significant challenges in this field.

Our research focuses on developing an end-to-end pipeline to accurately generate standard Bangla speech signals from Bangla dialect speech. Unlike previous studies that tackled separate tasks, we've taken an integrated approach by incorporating ASR, MT, and TTS into a single end-to-end framework for standardizing Bangla dialect text, illustrated in Fig.~\ref{fig:figure_1}. To address the scarcity of relevant datasets, we've collected raw dialect speech signals from the Noakhali region, which exhibits a distinct accent in the Bangla language due to cultural diversity. This study presents a groundbreaking method for utilizing multilingual language models (mLLMs) and ASR models to tackle the challenges of handling big data and performing ASR on dialects. Notably, prior work on Bangla ASR leveraging LLMs has been limited, making our approach potentially transformative for speech-related tasks in low-resource languages like Bangla.

Our contributions comprise \textit{\textbf{(1)}} development of a comprehensive dataset of specific Bengali dialects to facilitate the fine-tuning of multilingual ASR models for automatic dialect speech recognition. \textit{\textbf{(2)}} Annotation of dialect text to standard text for fine-tuning LLMs, enabling the MT of dialect text to standard text. \textit{\textbf{(3)}} Creation of a robust pipeline by integrating ASR, MT, and TTS into an efficient end-to-end framework for standardizing Noakhali speech signals into standard Bangla speech.

\section{Related works} 
\noindent \textbf{Dialect Speech Recognition:} 
Over the past decade, Bangladesh has made significant strides in speech-related studies. The field of ASR research took off in the early 1990s, yielding approximately 75 papers across various Bangla speech-related tasks \cite{sen2022bangla}. With the increasing relevance of Natural Language Processing (NLP), dialect recognition has emerged as a prominent area of interest in computer science. For instance, roughly 230 studies have been conducted on Chinese dialect recognition \cite{li2024chinese}. Similarly, machine learning and deep learning techniques have been utilized to develop ASR systems for diverse Arabic dialects \cite{9780142}. Notably, for low-resource languages like the Sudanese dialect, a recognition system achieved a label error rate (LER) of 73.67\% using a CNN model and a dataset of 7 hours and 50 minutes \cite{mansour2022end}. Furthermore, an innovative approach using DeepSpeech \cite{hannun2014deep} was employed to create an ASR system for the Tunisian dialect, achieving a word error rate (WER) of 24.4\% and a character error rate (CER) of 18.7\% with the TunSpeech dataset \cite{messaoudi2021tunisian}. A similar approach was taken by Pan et al. \cite{pan2019effective} to develop an end-to-end ASR system for the Lhasa dialect in Tibet, addressing limited data challenges with transformer-based models and unique acoustic modeling techniques. In Bangladesh, research on dialect recognition systems is gaining momentum. In 2023, Noor et al. \cite{noor2023real} developed a speech-to-text system using a word-matching algorithm for the Noakhali and Chittagong dialects, facing hurdles due to regional pronunciation variations and limited training data. Additionally, Sultana and Hossain \cite{sultanaHossain} utilized deep learning methods to classify Bangla dialects and gender across seven local languages. Recognizing the limitations of Bangla dialect datasets, in this investigation, we have constructed a Bangla dialect dataset and leveraged OpenAI's Whisper model dialect speech recognition. \\
\noindent\textbf{Bangla Language MT:}
The field of multilingual machine translation (MT) has significantly improved cross-language communication. Notably, recent efforts have involved translating Bangla to English using a sequence-to-sequence (seq2seq) model with an attention-based recurrent neural network (RNN) \cite{shiam2023neural}. While it is uncommon to translate regional dialects into standard Bangla\cite{faria2023vashantor} successfully implemented the BanglaT5 \cite{bhattacharjee2023banglanlg} and mT5 \cite{xue2020mt5} models, achieving a BLEU score of 69.06\% and a word error rate of 15.48\% for the Mymensingha dialect, as well as a 47.38\% BLEU score for the Noakhali dialect with 2500 samples. Milon et al.\cite{milon2020comprehensive} focused on converting the Chittagonian dialect using a word-to-word mapping technique and a multilingual lexicon. Given the minimal availability of the Noakhali dialect dataset in recent research, we incorporated a more extensive vocabulary from the Noakhali regions to develop a more robust translation model.

\noindent \textbf{Bangla Text-To-Speech System:}
The recent years have witnessed significant advancements in developing an efficient Bangla TTS system. Initially, Alam introduced a Bangla TTS system using the open-source Festival toolkit \cite{Alam2011}. Subsequently, in 2019, students from Shahjalal University of Science and Technology developed a Bangla TTS based on a Deep Neural Network, leveraging a 40 hours speech dataset comprising 12,500 utterances, which outperformed other existing systems at that time~\cite{9084052}. Building upon this progress, the same authors introduced the first end-to-end Bangla TTS system two years later, trained on 20 hours of speech data, which led to further performance enhancements \cite{bhattacharjee2021endendbanglaspeech}. While much attention has been given to the development of Bangla TTS systems, only a few researchers have focused on integrating TTS into dialect recognition systems in Bangla.  For instance, Begum et al. \cite{begum2019text} developed a TTS synthesis system for the Mymensingha dialect using the Festival toolkit, albeit encountering issues with rhythm in long sentences and a small dataset. The main limitations in current Bengali dialect speech recognition, translation, and TTS studies are the scarcity of available dialect-specific databases and the dependence on traditional methods, hindering comprehensive research in this field. In this study we constructed a large-scale Bengali dialect dataset, particularly focusing on the Noakhali dialect.  

\section{Methodology}
\begin{figure*}[t!]
\includegraphics[width=\textwidth]{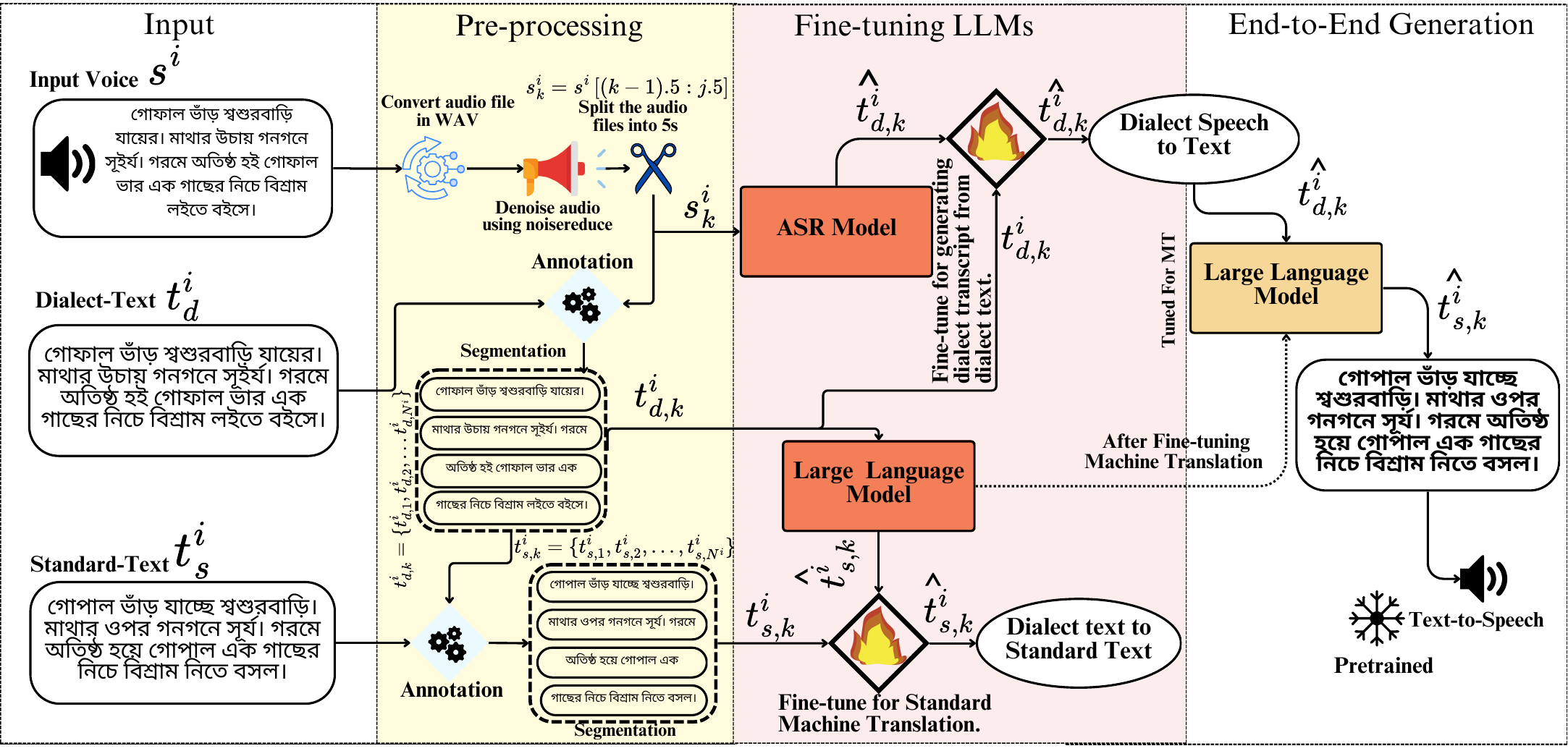}
\caption{\textit{BanglaDialecto} system: \textbf{(a)} input dialect speech signals \( s^i \) are converted into wav form, then it undergoes the process of noise reduction and splitting into manageable 5-second speech segments $s^i_k$, Dialect text $t^i_d$ and standard text $t^i_s$ then segmented in corresponding chunks $t^i_{d,k}$ and $t^i_{s,k}$. \textbf{(b)} The segment $s^i_k$ and $t^i_d$ are used to fine-tune the ASR to predict and transcript dialect speech $s^i_k$ into dialect text $t^i_d$. The other segment $t^i_{s,k}$ alongside with $t^i_{d,k}$ is used to fine-tune the LLMs for MT from dialect text to standard Bangla text.  \textbf{(c)} During end-to-end framework, the predicted transcript $\hat{t}^i_{d,k}$ of the dialect speech signal $s^i_k$ by the model $\mathcal{F}_1$ passes to the model $\mathcal{F}_2$ and then the model $\mathcal{F}_2$ predicts and translate standard Bangla text $\hat{t}^i_{s,k}$ from $\hat{t}^i_{d,k}$. We integrate a TTS model for generating standard Bangla speech signal from translated standard Bangla text $\hat{t}^i_{s,k}$.}
\label{fig:figure_2}
\end{figure*}
In this study, we intended to develop an end-to-end pipeline for converting Bengali dialect speech into the standardized Bangla speech signal. Prior research was done in this field separately focusing on either ASR or MT. We aim to integrate ASR, MT, and TTS for generating standard Bangla speech from dialectal Bangla speech signals.\\ 
\noindent\textbf{Problem Formulation:} Consider there are \( n \) number of dialect speech signals, where \( n \in N \). For each dialect speech signal, there is a corresponding dialect text, and considering the semantic structure of the Bengali dialect and standard form, for each dialect text, there is a unique standard text. Specifically, for the $i^{th}$ sample, the dialect speech signal is \( s^i \), the corresponding dialect text is \( t_{d}^i \), and the standard text is \( t_{s}^i \). We aim to build model $\mathcal{F}_1$ to transcript $s^i$ into corresponding \( t_{d}^i \) and model $\mathcal{F}_2$ to predict and generate \( t_{s}^i \) from \( t_{d}^i \). The objective is to build a two-stage model to convert dialect speech into standard text. This study focuses on accurately generating the standard text \( t^i_{s} \) from the dialect speech signal \( s^i \) leveraging LLMs. Then a TTS model will be used to complete the end-to-end pipeline through text-to-speech generation.

In this study, we focused on addressing two key challenges. \textbf{(a) Managing large-scale speech signals:}
Traditionally, Bangla ASR has relied on deep learning approaches. However, the emergence of LLMs in natural language processing has shown their potential in speech recognition for low-resource languages, as indicated by Liu et al.\cite{liu2024exploration} Although researchers are working to enhance the performance of low-resource ASR by leveraging transformer-based models, their application in Bangla speech-related tasks remains largely unexplored. A significant obstacle is the dearth of extensive and diverse audio datasets necessary for effectively fine-tuning ASR models for these tasks. The use of very small datasets has hindered research on the Bangla dialect, leading to less accurate performance \cite{noor2023real}. Our primary objective is to address these limitations by utilizing a large audio dataset that has not yet been utilized for Bangla dialect recognition or translation purposes. Through extensive fine-tuning of ASR and LLMs on speech datasets, our goal is to develop models capable of capturing linguistic variations. \textbf{(b) Improving performance in dialect translation:} LLMs are pretrained in standard Bangla language, making it challenging to capture dialectal vocabulary variations without extensive fine-tuning \cite{naveed2023comprehensive}. Modifying these models with large datasets covering syntax and vocabulary specific to dialects is crucial to produce accurate translation results. Previous research has primarily focused on building a bilingual dictionary for word-to-word mapping, which is time-consuming and less accurate when dealing with a larger lexicon \cite{milon2020comprehensive}. To address these issues, we aim to create a substantial translation dataset to overcome these shortcomings and enhance dialect translation accuracy. Fig.~\ref{fig:figure_2} depicts the architecture of our entire pipeline. 

\subsection{NDD: Noakhali Dialect Dataset}
\begin{figure}[t]
    \centering
    \includegraphics[width=1.0\linewidth]{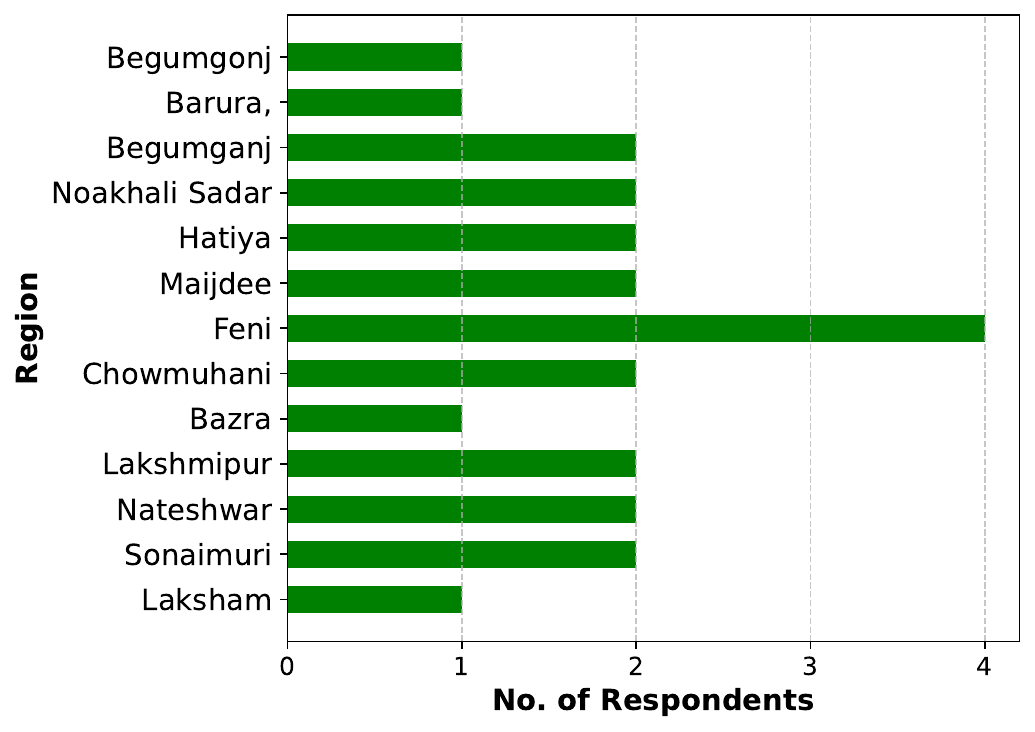}
    \caption{Interview participant distribution across the Noakhali region. Conducting interviews with respondents of different regions added diversity of speech accents to our NDD dataset.}
    \label{fig:figure_3}
\end{figure}

Understanding the various regional dialects of Bangladesh presents significant challenges, leading to a lack of available datasets. Data collection was therefore a critical and difficult aspect of this project. According to the literature, no speech signal dataset for the Noakhali dialect has been available to date. So, we manually gathered accented audio samples from diverse sources, including YouTube videos and Facebook groups shared among native speakers, and conducted interviews with residents in Noakhali, Bangladesh. Participants were asked to read a standard paragraph in their native accent, and these recordings were used to develop our dataset. In total, we collected 10 hours of Noakhali regional speech data, denoted as $s^i$. The Noakhali Dialect Dataset, \textbf{NDD} comprises three columns: Noakhali regional speech data, $s^i$; each speech signal's $s^i$ annotated dialect text version, $t^i_d$; and standard Bangla translation form $t^i_s$ of each dialectal text, $t^i_d$. The annotation is conducted by highly qualified human evaluators from Noakhali. We gathered data from 24 humans, 18 of whom were men and 6 of whom were women. Mostly, they are aged between 18 and 28. Demographic details are shown in table~\ref{tab:combined_reordered}. In Fig.~\ref{fig:figure_3}, the locations of Noakhali, from where we collected data have been shown.  This constructed NDD dataset then undergoes preprocessing to handle big data management challenges. The following preprocessing NDD dataset is utilized for fine-tuning the ASR model and LLM for the dialect speech-to-dialect text transcription and dialect text-to-standard text translation tasks. 
\subsection{Preprocessing: Denoising and Splitting Voice Data}
Before preparing for fine-tuning ASR, each dialect speech signal, $S^i$ was denoised to remove background noise and improve clarity, ensuring higher-quality inputs for the model. This step is essential for accurate transcription and translation. 
\begin{table}[!t]
\centering
\caption{Detailed statistics of the actual NDD dataset, along with the demographic profile of interview participants who volunteered to provide dialect speech audio. We conducted interviews taking their consent, and asked them to read our standard Bangla text in their Noakhali accent. We recorded that and proceeded to NDD dataset construction.}
\label{tab:combined_reordered}
\setlength{\tabcolsep}{0.80em} 
\renewcommand{\arraystretch}{1} 
\scalebox{0.8}{ 
\begin{tabular}{c}
\toprule
\textbf{NDD Dataset Details} \\
\midrule
\begin{tabular}{@{}l|l@{}}
\textbf{Characteristic} & \textbf{Value} \\ \hline
Unique Characters & 72 \\ 
Max Text & \includegraphics[trim={1cm 9.7cm 0cm 9.8cm},clip,width=0.8\linewidth]{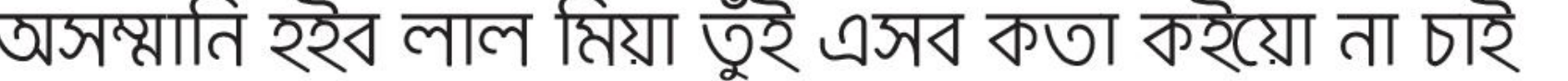} \\ 
Total Unique Words & 11246 \\ 
Audio Duration & 10 hours \\ 
Audio Sample Rate & 16 kHz \\ 
Bit Rate & 16 \\ 
Single Channel & Mono \\ 
\end{tabular} \\
\midrule[\heavyrulewidth]
\textbf{Demographic Data of Interview Participants} \\
\midrule
\begin{tabular}{@{}cc|cc@{}}
\textbf{Characteristics} & \textbf{Category} & \textbf{Number} & \textbf{Percentage} \\ \hline 
\multirow{3}{*}{\textbf{Age}} & 18 to 28 & 20 & 83.33\% \\
 & 29 to 38 & 2 & 8.33\% \\
 & 39 to 48 & 2 & 8.33\% \\ \hline
\multirow{2}{*}{\textbf{Gender}} & Male & 18 & 75\% \\
 & Female & 6 & 25\% \\ 
\end{tabular} \\
\bottomrule
\end{tabular}}
\end{table}
\noindent \textbf{Managing Long Speech Data: }To manage long speech signals, each denoised signal \( s^i \) was split into smaller, 5-second segments. Given a speech signal \( s^i \) of total duration \( T^i \), it was divided into \( N^i \) segments:

\[
N^i = \left\lfloor \frac{T^i}{5} \right\rfloor
\]

Each segment \( s^i_k \) (where \( k = 1, 2, \dots, N^i \)) represents a 5-second portion of \( s^i \), defined as: $s^i_k = s^i \left[ (k-1) \cdot 5 : j \cdot 5 \right]$
If the total duration \( T^i \) is not a multiple of 5, the remaining portion of the signal, \( s^i_{\text{remainder}} \), is truncated to fit the architecture of transformer models. This ensures that all input segments are of equal length, aligning with the fixed-length input requirement of transformers.

\noindent \textbf{Text Splitting Mechanism: }
After splitting the dialect speech signal \( s^i \) into 5-second segments, the corresponding dialect text \( t^i_d \) and standard text \( t^i_s \) were also split to align with the speech segments. The text splitting follows the same segmentation process to ensure that each chunk of \( s^i \) can be fine-tuned with its corresponding text segment. For the \( i^{th} \) data sample, the dialect text \( t^i_d \) and the standard text \( t^i_s \) are split into \( N^i \) segments:
\[
t^i_d = \{ t^i_{d,1}, t^i_{d,2}, \dots, t^i_{d,N^i} \}
\]
\[
t^i_s = \{ t^i_{s,1}, t^i_{s,2}, \dots, t^i_{s,N^i} \}
\]
Each segment \( t^i_{d,k} \) and \( t^i_{s,k} \) corresponds to the dialect and standard text that matches the speech segment \( s^i_k \), ensuring synchronization. This alignment allows each 5-second chunk of the speech signal \( s^i_k \) to be fine-tuned with the corresponding chunk of dialect text \( t^i_{d,k} \) during the first-stage fine-tuning, and with the standard text \( t^i_{s,k} \) during the second stage.

This process ensures coherent and accurate training across both the speech-to-text and text-to-text transformations, matching the segmented inputs.
\subsection{Employing ASR and LLM in dialect standardization} 
We employ ASR and LLM in two scenarios: \textbf{(a)} converting dialect speech to dialect text and \textbf{(b)} generating standard text from dialect text. 

\begin{table*}[h]
\small
\centering
\caption{Comparing model performances with other approaches that were used for dialect transcription and translation after fine-tuning the ASR and LLMs on the NDD dataset. The outcomes demonstrate that our models performed better than any of the current techniques. Research on the Noakhali dialect is still lacking in both MT and ASR. For this reason, we conducted a comparison between our performance and other languages' dialect analytics. $\uparrow$ ($\downarrow$) means higher (lower) is better. ‘-’ denotes results that are not available there. }
\label{tab:res_comp}
\setlength{\tabcolsep}{0.2em} 
\renewcommand{\arraystretch}{1.2} 
\begin{tabular}{ccccccccc}
\hline
\multicolumn{3}{c}{\textbf{ASR }} &
  \multicolumn{4}{c}{\textbf{MT}} &
   &
   \\ \cmidrule(lr){1-4} \cmidrule(lr){5-9} 
\textbf{Methods} &
  \textbf{Language} &
  \textbf{CER(\%)$\downarrow$} &
  \textbf{WER(\%)$\downarrow$} &
  \textbf{Methods} &
  \textbf{Language} &
  \textbf{CER(\%)$\downarrow$} &
  \textbf{WER(\%)$\downarrow$} &
  \textbf{BLEU(\%)$\uparrow$} \\ \hline

Messaoudi el al.\cite{messaoudi2021tunisian} &
  Tunisian dialect &
  18.7 &
  24.4 &
  Faria et al.\cite{faria2023vashantor} &
  Mymensingh dialect &
  8.23 &
  15.48 &
  \textbf{69.06} \\
Ying et al.\cite{ying2020sichuan} &
  Sichuan dialect &
  25.24 &
  57.02 &
  Faria et al.\cite{faria2023vashantor} &
  Noakhali dialect &
  20.3 &
  38.7 &
  47.43 \\
Alam el al\cite{alam2022bengali} &
  Standard Bangla  &
  13.856  &
  39.29 &
  Kchaou et al.\cite{kchaou2023hybrid} &
  Tunisian dialect &
  - &
  - &
  60.00 \\
 Kabir et al.\cite{kabir2024automatic} &
  Sylhet Dialect &
  - &
  14.89 &
Hamed et al.\cite{hamed2021deep} &
  Arabic &
   - &
   -&
  18 \\
Nasr et al.\cite{nasr2023end} &
  Arabian dialect &
  20 &
  30 &
  Baruah et al.\cite{baruah2021low} &
  Assamese &
  - &
  - &
  50.19 \\
Phung et al.\cite{phung2024enhancing} &
  Vietnamese &
  4.79 &
  2.79 &
  Adiputra et al.\cite{sulaeman2015development} &
  Japanese-Indonesian &
  - &
  - &
  8.78 \\
Safieh et al.\cite{safieh2022end} &
  Jordanian dialect &
  26.4 &
  51.50 &
  Prama et al.\cite{pramasylheti} &
  Sylhet Dialect &
  - &
  - &
  57.4 \\
Ours &
  Noakhali Dialect &
  \textbf{0.8} &
  \textbf{1.5} &
  Ours &
  Noakhali dialect &
  \textbf{20.2} &
  \textbf{38.2} &
   41.6 \\ \hline
\end{tabular}
\end{table*}

\noindent \textbf{(a) Convert Dialect Speech to Dialect Text: } 
Fine-tuning is essential for adapting pre-trained models to specialized tasks, improving relevance and accuracy. In this study, we fine-tuned models for both speech-to-text conversion and translation. Pre-trained models often struggle with dialect recognition and nuanced language variations. By fine-tuning, we customized the models to capture the linguistic features of the Noakhali dialect for improved speech-to-text accuracy. Whisper ASR models—base (74M), small (244M), medium (769M), and large (1550M)—were fine-tuned on the \textbf{NDD} dataset. Each speech segment $s^i_k$ was paired with its corresponding dialect text $t^i_{d,k}$, forming the input-output pair ($s^i_k$, $t^i_{d,k}$). The objective of fine-tuning is to generate accurate $t^i_{d,k}$ from $s^i_k$ through LLMs.  

\noindent \textbf{(b) Text standardizing from dialect: }To generate standard Bangla text from dialect text, we fine-tuned several models: XLM-ProphetNet (616M parameters), mBART50 (611M), IndicBART (244M), mT5 (base, 582M), and BanglaT5 (247M). Fine-tuning these models on the \textbf{NDD} dataset enabled them to adapt to dialect-specific linguistic features, ensuring more accurate translation into standard Bangla $t^i_{s,k}$. The dialect text $t^i_{d,k}$ served as the input, and the corresponding standard Bangla text $t^i_{s,k}$ as the target output, improving the models’ ability to standardize text while preserving linguistic coherence.

\noindent \textbf{Model Architecture: } We utilized three different models for our study. \textbf{(a)} Some studies have already been done in low-resource languages, and authors got better performance by fine-tuning the whisper ASR model. They encourage us to use this whisper ASR model in our study to see how whisper fine-tuning works in the Bangla dialect. Whisper~\cite{radford2022robustspeechrecognitionlargescale} is a powerful speech recognition system created by OpenAi. It utilized an encoder-decoder transformer architecture to transcribe and translate speech in many languages effectively. Whisper, trained on a large dataset of 680,000 hours of multilingual audio, can generalize across varied linguistic contexts and acoustic surroundings. It has 5 versions, tiny with the lowest parameters of 39M and large with 1550M parameters. Whisper implements the GeLU\cite{hendrycks2016gaussian} activation mechanism in its Transformer-based architecture. \textbf{(b)} The BanglaT5 model is a sequence-to-sequence transformer model for the Bengali language, which was pre-trained on a large clean corpus of 27.5 GB of Bangla data. It is the basic T5 model, including 12 layers, 12 attention heads,  768 hidden size, a feed-forward size of 2048. In feed-forward layer they utilized the GeGLU activation function\cite{shazeer2020glu}  \textbf{(c)}  
AlignTTS~\cite{zeng2020aligntts} is used for generating standard audio speech signals from the standard text. AlignTTS is a speech synthesis model that predicts mel-spectrums from text using a Feed-Forward Transformer. It replaces the attention mechanism with alignment loss and dynamic programming for efficient text-to-speech alignment.  

\section{Experiment}

\subsection{Setup}
\noindent\textbf{Dataset:}
We fine-tuned the ASR and LLMs using our constructed dataset, \textbf{NDD}. After applying the pre-processing strategy on big data $s^i$, $t_d^i$ and $t_s^d$, and the data samples turned to manageable $s^i_k$, $t^i_{d,j}$ and $t^i_{s,k}$ segments. After segmentation, the NDD dataset comprises 7200 sample data points. We partitioned this dataset into training, validation, and testing sets. The training set consists of 6270 samples, the validation set contains 810 samples, and the test set includes 120 samples. This division ensures that the models are trained on a substantial portion of the data, with a smaller portion reserved for validation and testing, which is a standard and thereby enables rigorous evaluation of model performance.\\
\noindent\textbf{Implementation Details
\footnote{Code and data: \url{https://github.com/EncryptedBinary/BanglaDialecto}}
:}
\begin{table*}[t]
\small
\centering
\caption{Comparative evaluation of pretrained and fine-tuned models on dialect speech-to-text and dialect text to standard text translation tasks. The models are assessed using three metrics: character error rate (CER), word error rate (WER), and BLEU score. For STT, Whisper-large V2 shows the best results with CER of 0.8\% and WER of 1.5\%. For DTT, BanglaT5 outperforms others with a BLEU score of 41.6, demonstrating superior performance in translation tasks. Fine-tuning consistently improves model accuracy across all metrics.}
\label{tab:llm_comp}
\setlength{\tabcolsep}{0.015em} 
\renewcommand{\arraystretch}{1.3} 
\begin{tabular}{ccccccccc}
\hline
\multirow{2}{*}{Task} &
  \multirow{2}{*}{Models} &
  \multirow{2}{*}{Parameter} &
  \multicolumn{3}{c}{\textbf{Pre Trained version}} &
  \multicolumn{3}{c}{\textbf{Fine-tuned version}} \\ \cmidrule(lr){4-6} \cmidrule(lr){7-9} 
 &
   &
   &
  \textbf{CER(\%)$\downarrow$} &
  \textbf{WER(\%)$\downarrow$} &
  \textbf{BLUE Score(\%)$\uparrow$} &
  \textbf{CER(\%)$\downarrow$} &
  \textbf{WER(\%)$\downarrow$} &
  \textbf{BLUE Score(\%)$\uparrow$} \\ \hline
\multirow{5}{*}{\begin{tabular}[c]{@{}c@{}}Dialect Speech-to-text\\ (ASR)\end{tabular}} &
  Sequential Transformer &
  24M &
  87 &
  179.8 &
  - &
  \multicolumn{1}{l}{} &
  \multicolumn{1}{l}{} &
  \multicolumn{1}{l}{} \\
 & Whisper-base     & 74M   & 230.8 & 232.2 & - & 20.6 & 47.1  & -    \\
 & Whisper-small    & 244M  & 389.8 & 411.9 & - & 2.0  & 4.0   & -    \\
 & Whisper-medium   & 769M  & 259.2 & 169.0 & - & 1.5  & 2.0   & -    \\
 & \textbf{Whisper-large V2} & 1550M & 135.2 & 167.5 & - & \textbf{0.8}  & \textbf{1.5}   & -    \\ \hline
\multirow{4}{*}{\begin{tabular}[c]{@{}c@{}}Dialect text Translation\\ (MT)\end{tabular}} &
  mBART50 &
  611M &
  248.5 &
  612.8 &
  0.45 &
  79.8 &
  416.8 &
  3.0 \\
 & IndicBART        & 244M  & 166.5     & 409.8     & 11.9 & 22.8 & 118.6 & 27.7 \\
 & mT5(base)        & 582M  & 275.4     & 405.4     & 5.6 & 34.4 & 59.2  & 18.6 \\
 & \textbf{BanglaT5}         & 247M  & 178.9     &  281.6     & 22.7 & \textbf{21.3} & \textbf{38.2}  & \textbf{41.6} \\ \hline
\end{tabular}
\end{table*}
Before proceeding towards fine-tuning ASR, \textbf{(a)} we used \textit{noisereduce library} \footnote{noisereduce: \url{https://pypi.org/project/noisereduce/}} to reduce the noise in \textit{wav} form audio signal. Then \textit{pydub library}\footnote{pydub: \url{https://pypi.org/project/pydub/}} was employed to partition the big audio data $s^i$ into smaller manageable segments, $s^i_k$. Annotation was done manually on these dialect audio signals, $s^i_k$, with dialect transcription $t^i_d,k$, and then, in a similar way, dialect text $t^i_d,k$ was translated to standard Bangla text $t^i_s,k$. Supervised fine-tuning was applied considering $s^i_k$ as input and $t^i_d,k$ as the target on different variants of Whisper models for the Noakhali speech recognition. Similarly, for MT, different LLMs are fine-tuned using $t^i_d,k$ as input and the target, $t^i_s,k$. \textbf{(b)}  We use a whisper feature extractor, whisper tokenization, and whisper processor for Dialect transcription. For BanglaT5, we use \textit{Autotokenizer} for the tokenizer. All models are trained for 10 epochs, 16 batches for ASR, 25 epochs, and 6 batches for translation. All experiments were conducted on the NVIDIA Tesla A100 GPU using the \textit{PyTorch} framework. \textbf{(c)} After the generation of standard Bangla text, AlignTTS is used for producing the standard Bangla speech signal.\\
\noindent \textbf{Evaluation metrics: }In the evaluation process, two steps are involved: one for assessing the performance of transcript dialect voice to dialect text and another for translating dialect text to standard text. For evaluation purposes, the following metrics are utilized:
\textit{\textbf{(a) Character error rate(CER):}} CER\cite{wobbrock2006analyzing} is an evaluation metric used to assess the accuracy of speech recognition and translation systems by quantifying the number of character-level errors, including insertions, deletions, and substitutions in the machine-generated text as compared to the reference text. A lower percentage indicates better performance in the generated text. \textit{\textbf{(b) Word error rate(WER)}:}WER\cite{zechner2000minimizing} is a widely used metric to gauge the model performance on speech recognition or machine translation systems. It computes the disparity between the generated output and the reference text in terms of words. It evaluates the performance of the generated text by dividing the errors, such as substitutions, deletions, and insertions, by the total number of words in the reference text. A lower WER signifies that the generated output is closer to the reference text. \textit{\textbf{(c) Bilingual Evaluation Understudy(BLEU):}} BLEU\cite{papineni2002bleu}, introduced by IBM in 2001, was one of the pioneering metrics for assessing the performance of machine-generated text, particularly in machine translation tasks. It utilizes the n-grams approach to evaluate the quality of the output text by analyzing the similarity of the generated text to the reference text.

\subsection{Results and Analysis}
In this section, we present the results of our novel approach for converting Noakhali dialect speech into standard Bangla using an integrated, end-to-end framework. Our system combines ASR with OpenAI’s fine-tuned Whisper models, MT using BanglaT5, and text-to-speech (TTS) via AlignTTS. To the best of our knowledge, no previous work has integrated these three distinct speech-related tasks into a unified framework. Furthermore, we utilized a large-scale audio and text dataset for fine-tuning the multilingual ASR and LLMs, which is uncommon for Bangla dialects. For comparison, Table~\ref{tab:res_comp} presents evaluation results from previous studies, while Table~\ref{tab:llm_comp} illustrates the performance improvements observed in our study. 
\begin{table*}[!t]
    \centering
    
    \caption{End-to-end generation from our fine-tuned models. For each input dialect voice, the model Whisper-large V2 generated the dialect texts, listed in the ASR column, from which the BanglaT5 model generated standard Bangla Text.}
    \begin{tabularx}{\textwidth}{X X X}
        \hline
        Input Voice & ASR (Whisper Large V2) & MT (BanglaT5) \\ \hline
        \includegraphics[trim={0cm 3.2cm 0cm 3.2cm},clip,width=\linewidth]{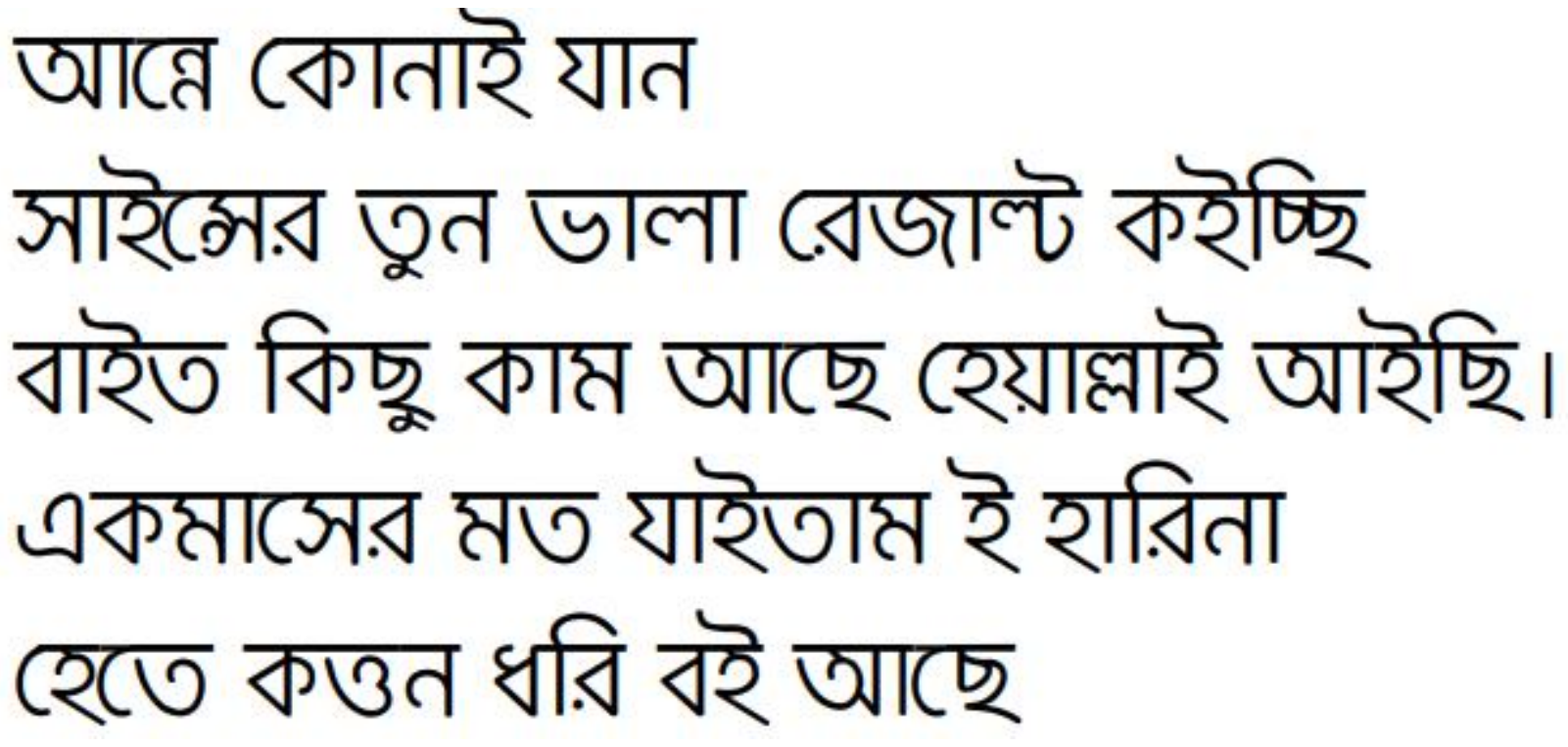} & 
        \includegraphics[trim={0cm 3.2cm 0cm 3.2cm},clip,width=\linewidth]{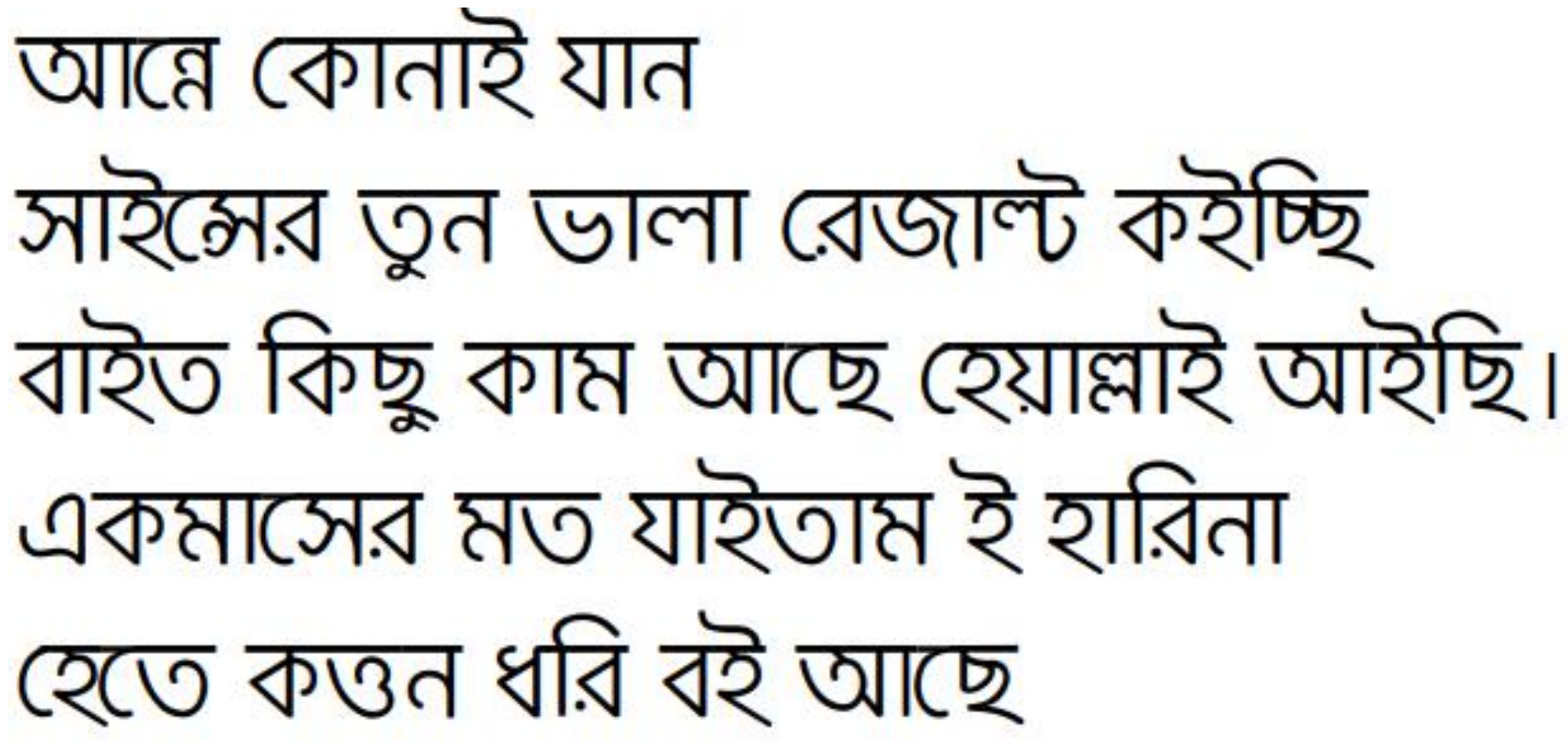} & 
        \includegraphics[trim={0cm 3.2cm 0cm 3.2cm},clip,width=\linewidth]{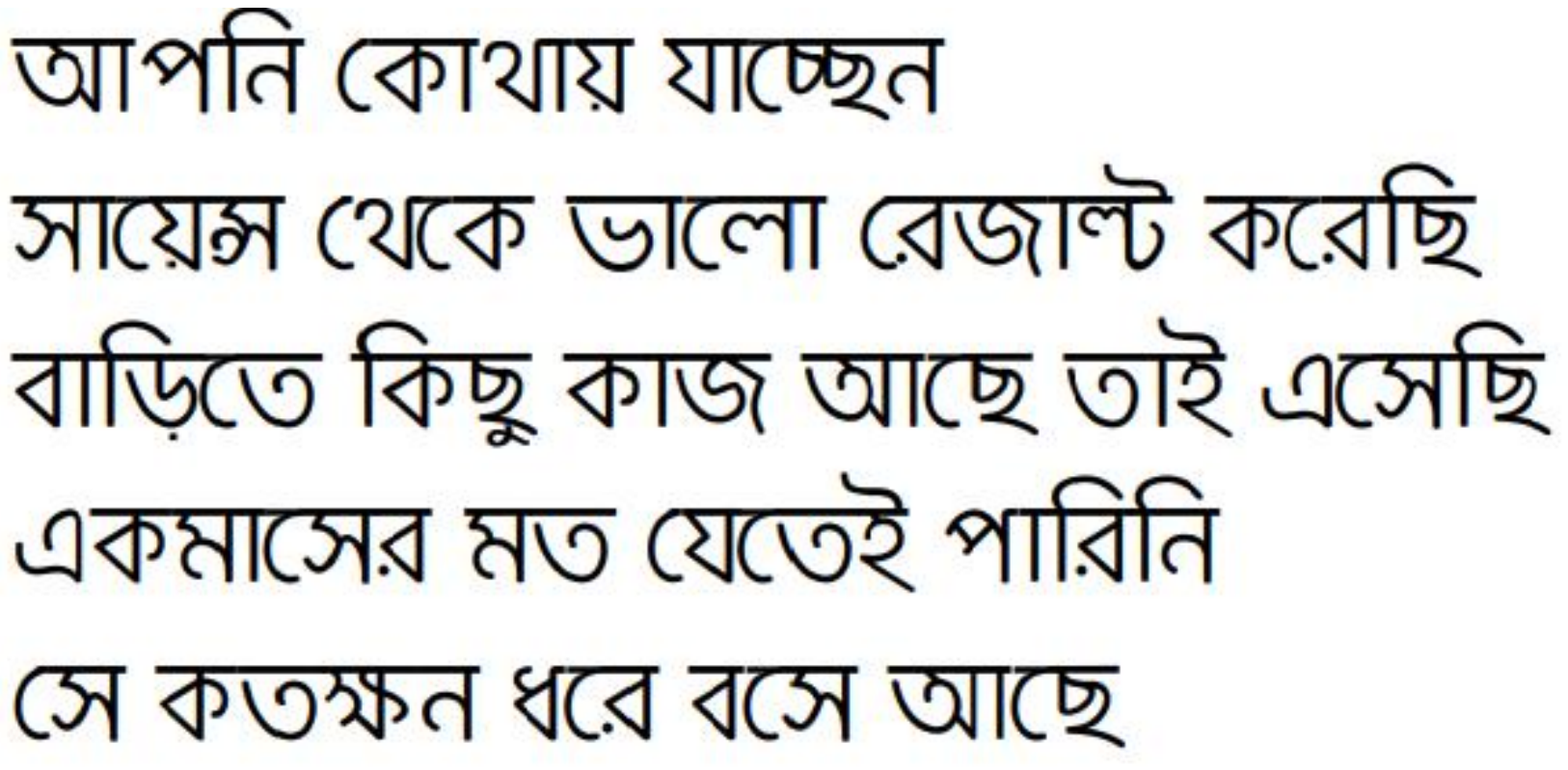} \\ 
        
        \hline
    \end{tabularx}
    \label{tab:exam}
\end{table*}

\begin{table*}[h]
    \small
    \centering
    \caption{English pronunciations of dialect speech examples given in Table~\ref{tab:exam}. The ``Input Voice" column presents the dialect pronunciation, ``ASR" shows the transcribed text, ``BanglaT5" lists the standard Bangla translation, and the fourth column provides the English translation of the Bangla text.}
    \setlength{\tabcolsep}{0.008em}
    \renewcommand{\arraystretch}{1.2} 
    \begin{tabularx}{\textwidth}{p{4.2cm} p{4.2cm} p{4.5cm} p{5cm}} 
        \hline
        Input Voice & ASR (Whisper Large V2) & MT (BanglaT5) & English (Translation)  \\ \hline
        Anne konai jan & Anne konai jan & Apne kothay jan & Where are you going \\
        Science tun vala result koichi & Science tun vala result koichi & Science theke valo result korechi & Have done a good result from Science \\
        Bait kichu kam ase hellai aichi & Bait kichu kam ase hellai aichi & Barite kichu kaj ase tai esechi & Have some work at home so I came \\
        Ekmasher moto jaitami harina & Ekmasher moto jaitami harina & Ekmasher moto jatei parina & Can't go since one month \\
        Hete koto khn dhori boi ase & Hete koto khn dhori boi ase & She koto kokhn dhore bose ache & How long has he been sitting \\
        \hline
    \end{tabularx}
    \label{tab:sample}

\end{table*}

\noindent \textbf{Bangla Dialect Speech Recognition:}
At the very beginning, we tried different audio models to check the performance of our custom audio dataset. Nowadays LLMs are becoming a source of speech-related tasks\cite{latif2023sparks}. Finally, We have seen the best output is provided by Open AI’s whisper large model. We fine-tuned all of the available Whisper models with our constructed NDD dataset. The comparative analysis between pre-trained and fine-tuned Whisper models reveals significant performance improvements in the fine-tuned variants, which is illustrated in Table-~\ref{tab:llm_comp}. Specifically, for the Dialect Speech-to-Text (STT) task, the Whisper-large V2 model, when fine-tuned, demonstrates a substantial reduction in Character Error Rate (CER) and Word Error Rate (WER) compared to its pre-trained counterpart. The pre-trained Whisper-large V2 model exhibits a CER of 135.2\% and a WER of 167.5\%, while the fine-tuned version achieves a remarkable CER of 0.8\% and a WER of 1.5\%. This reduction underscores the efficacy of fine-tuning in enhancing the model's accuracy and efficiency in transcribing dialectal speech, highlighting its superior capability to handle complex linguistic variations with greater precision. In contrast, the pre-trained Whisper models, including Whisper-base, Whisper-small, and Whisper-medium, exhibit higher CER and WER values, indicating their limited effectiveness in addressing dialectal speech nuances without fine-tuning. The improvement in fine-tuned Whisper models demonstrates their enhanced adaptability and performance in specialized tasks, making them significantly more effective for practical applications in speech-to-text technology. The fine-tuning process not only improves the accuracy of these models but also emphasizes their potential for achieving higher efficiency and impact in real-world scenarios.

\noindent \textbf{Performance Analysis of Translation Models: }In the Dialect Text-to-Translation (DTT) task, fine-tuning notably enhances model performance. The pretrained models, such as mBART50, IndicBART, and mT5 (base), exhibit comparatively lower BLEU scores, with mBART50 achieving a BLEU score of 0.45\% and IndicBART and mT5 (base) scoring 11.9\% and 5.6\%, respectively. In contrast, the fine-tuned BanglaT5 model achieves a BLEU score of 41.6\%, the highest among the evaluated models. This significant improvement demonstrates the efficacy of fine-tuning in refining translation capabilities, particularly for dialect-specific tasks. The fine-tuned BanglaT5 model's superior BLEU score highlights its enhanced accuracy and fluency in translation, making it more effective for handling nuanced dialectal variations and providing more reliable translations, indicating in Table~\ref{tab:exam}. The results underscore the impact of fine-tuning in optimizing model performance and achieving higher translation quality. 

Our observations include \textbf{\textit{(a)}} \textcolor{black}{fine-tuning Whisper multilingual models }and MT models on a large-scale speech signal dataset, NDD allows them to adapt the complexity of both dialect speech to dialect text and dialect text to standard text tasks, \textbf{\textit{(2)}} fine-tuned models excel in handling dialectal variations, delivering superior results in both speech-to-text and translation tasks, Whisper-large V2’s CER and WER drop dramatically, and BanglaT5 achieves the highest BLEU score among models, and \textbf{\textit{(c)}} after completing the end-to-end pipeline, the AlignTTS handle the standardized speech, indicating real-life reliability of our method.

\subsection{Ablation Study}
We conduct our ablation studies on our two main tasks. One is the variations according to the parameter enhancement on Dialect transcription. The other variation is according to the models. This study gives a new insight into the performance of the project.


%

\begin{figure}[!t]
    \centering
    \includegraphics[scale=0.67]{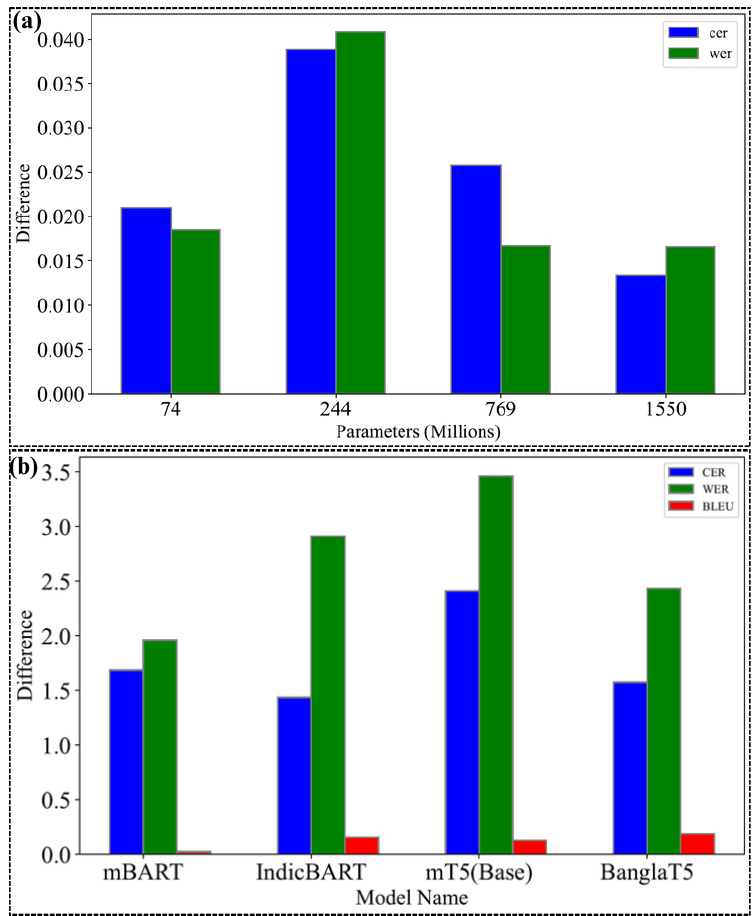}
\caption{(a) The impact of increasing Whisper model parameters on accuracy, as measured by CER and WER. It shows that larger models, particularly Whisper-large V2, yield better performance, even in their pre-trained state, due to their higher parameter count. (b)  Comparison of mBART, IndicBART, mT5 (Base), and BanglaT5 across CER, WER, and BLEU metrics. mT5 (Base) has the highest CER, mBART performs best in WER, and BLEU scores are similar for all models.}
\label{fig:combined}
\end{figure}

\noindent \textbf{Variations of Model Parameter Dialect Transcription:} We experimented with multiple versions of whisper models to identify differences in the results. We obtained insightful results by calculating the change rate between pre-trained and fine-tuned models for each whisper model using CER and WER. The precision of the results improves as the parameters of the Whisper model increase, as depicted in figure \ref{fig:combined}(a). Notably, \textit{whisper-large-V2} yields superior results, even in its pre-trained state, owing to its high parameter count.

\noindent
\textbf{Variations of Models' architecture on MT:} 
Fig.~\ref{fig:combined}(b) compares the performance of four models—mBART, IndicBART, mT5 (Base), and BanglaT5—across CER, WER, and BLEU. mT5 (Base) shows the highest CER, indicating more character-level errors, while BanglaT5 and IndicBART perform better. For WER, IndicBART and mT5 (Base) have higher errors, with mBART showing the lowest WER. BLEU scores are similar across all models, reflecting comparable fluency and correctness.

\subsection{Discussion}
This project aims to construct an end-to-end pipeline for generating standard Bangla speech from regional Bangla dialect speech. For investigation, we considered the Noakhali dialect due to the diverse accent. \textcolor{black}{We fine-tuned the Whisper ASR model} for the task of transcribing the Noakhali dialect speech signal to the Noakhali dialect text. Then for the next task, we fine-tuned LLMs for translating the Noakhali dialect text to standard Bangla text. To wrap the pipeline we used the open sourced TTS model for generating predicted standard Bangla speech signals. Our results demonstrate the effectiveness of fine-tuning multilingual ASR and LLMs for dialectal speech recognition and translation. Fine-tuning significantly improved the accuracy of Whisper models for speech-to-text conversion, particularly Whisper-large V2, which achieved remarkable reductions in CER and WER. This indicates its ability to handle complex dialectal speech with precision. Similarly, for text-to-text translation, the fine-tuned BanglaT5 model outperformed others, achieving the highest BLEU score, showcasing its superiority in translating dialect text to standard Bangla. These improvements highlight the potential of LLMs to tackle linguistic diversity effectively, ensuring higher performance across dialect-to-standard text tasks. Our approach, integrating ASR, MT, and TTS, provides a robust pipeline for practical applications in Bangla dialect standardization.  

\noindent\textbf{Limitations:} 
Despite promising results, the study is limited by its focus on the Noakhali dialect, which restricts generalizability across the 55 Bangla dialects. The NDD dataset from specific regions may not fully represent dialectal diversity, affecting model reliability. Expanding the dataset and incorporating more dialects is needed for broader applicability.

\section{Conclusion}
The project mainly focuses on the concept of Bangladeshi dialect recognition along with converting various Bengali accents into a standardized, formal Bengali voice. The technology recognizes regional sound differences in spoken Bengali and converts them into a clear, neutral form using powerful speech processing technologies. 
The main goal of this project is to enhance communication in formal contexts, such as medical purposes, media, and education, where a formal language is favored. This project will encourage native people from different regions to participate and experience new things previously difficult for them due to language barriers. This study successfully demonstrates the construction of an end-to-end pipeline for converting Noakhali dialect speech into standard Bangla speech using fine-tuned LLMs.  The investigation introduces a large-scale dialectal Noakhali speech signal dataset, NDD. By leveraging Whisper for speech recognition, BanglaT5 for translation, and AlignTTS for speech synthesis, we achieved significant improvements in transcription and translation accuracy. Whisper-large V2 reduced CER and WER, while BanglaT5 delivered high BLEU scores, highlighting the effectiveness of fine-tuning in handling dialectal variations. Regional voice to standard Bangla voice can be a great prospect for our country. People from different regional districts can use it in every aspect of their life. Although focused on the Noakhali dialect, this approach lays the foundation for future research into Bangla dialect standardization, with potential applications across various dialects.

\noindent\textbf{Future works:} 
Future improvements should include increasing dataset diversity by incorporating more regional accents and expanding the system’s multilingual capabilities to support communication across different languages. Enhancing contextual understanding of dialects can further refine dialect conversion processes, potentially benefiting educational, entertainment, and medical applications for local communities.

\section*{{Ethics Statement}}
This study was conducted with a firm commitment to ethical standards, particularly in the development and assessment of dialect speech recognition using LLMs within a scientific framework. We reached out to a total of 24 participants from diverse regions of Noakhali, ensuring that participation was voluntary and based on informed consent. The majority of participants were students, many of whom had recently graduated. In order to safeguard the safety and well-being of all individuals involved, significant measures were implemented to minimize any potential physical or psychological risks during the evaluation process. Recognizing the substantial impacts that even minor errors in scientific research can have, we implemented stringent procedures to ensure that all content produced adhered to the highest ethical standards. Our methodology was designed to avoid language or findings that could potentially perpetuate harm or inequality based on race, gender, or other social determinants of health. By upholding these principles, our study was conducted with the utmost ethical integrity, fostering a respectful environment for collaboration between our work and the community at large.




\bibliographystyle{IEEEtran}
\bibliography{reference}

\end{document}